\documentclass[runningheads]{llncs}

\usepackage{eccv}

\usepackage{eccvabbrv}

\usepackage{graphicx}
\usepackage{booktabs}
\usepackage{footnote}

\usepackage[accsupp]{axessibility}  

\usepackage{hyperref}

\usepackage{orcidlink}
\usepackage{multirow} 
\usepackage{bbding}
\usepackage{color}

\newcommand{\syx}[1]{{\color{black} #1}}

\begin{document}

\title{Multi-Task Domain Adaptation for Language Grounding with 3D Objects} 

\titlerunning{Multi-Task Domain Adaptation for Language Grounding with 3D Objects}


\author{Penglei Sun\inst{1}\thanks{Equal Contribution.}\and
Yaoxian Song\inst{2}$^{\star}$ \and
Xinglin Pan\inst{1} \and 
Peijie Dong\inst{1} \and 
Xiaofei Yang\inst{13} \and \\
Qiang Wang\inst{4}\textsuperscript{\Envelope} \and 
Zhixu Li\inst{5} \and 
Tiefeng Li\inst{2} \and
Xiaowen Chu\inst{1}\textsuperscript{\Envelope}
}

\authorrunning{P. Sun et al.}

\institute{
The Hong Kong University of Science and Technology (Guangzhou), China 
\email{\{psun012,xpan413,pdong212\}@connect.hkust-gz.edu.cn},
\email{xwchu@ust.hk}
\and
Zhejiang University, China 
\email{\{songyaoxian,litiefeng\}@zju.edu.cn}\\
\and
Guangzhou University, China 
\email{xiaofeiyang@gzhu.edu.cn}
\and
Harbin Institute of Technology (Shenzhen), China
\email{qiang.wang@hit.edu.cn}
\and
Fudan University, China
\email{zhixuli@fudan.edu.cn}
}

\maketitle

\begin{abstract}
The existing works on object-level language grounding with 3D objects mostly focus on improving performance by utilizing the off-the-shelf pre-trained models to capture features, such as viewpoint selection or geometric priors. 
However, they have failed to consider exploring the cross-modal representation of language-vision alignment in the cross-domain field. 
To answer this problem, we propose a novel method called \textbf{D}omain \textbf{A}daptation for \textbf{L}anguage \textbf{G}rounding (\textbf{DA4LG}) with 3D objects.
Specifically, the proposed DA4LG consists of a visual adapter module with multi-task learning to realize vision-language alignment by comprehensive multimodal feature representation.
Experimental results demonstrate that DA4LG competitively performs across visual and non-visual language descriptions, independent of the completeness of observation. 
DA4LG achieves \textbf{state-of-the-art} performance in the single-view setting and multi-view setting with the accuracy of \textbf{$83.8 \%$} and \textbf{$86.8 \%$} respectively in the language grounding benchmark SNARE.
The simulation experiments show the well-practical and generalized performance of DA4LG compared to the existing methods.
Our project is available at \url{https://sites.google.com/view/da4lg}.

\end{abstract}

\section{Introduction}
\label{sec:intro}

Visual language grounding aims to identify the region or object within visual content described by natural language~\cite{chen2020scanrefer,harnad1990symbol}. 
It serves as an essential bridge for current embodied agents to connect symbolic concepts with the perceptible real world, enabling the evolution of an agent's intelligence from perceptual to cognitive decision-making~\cite{bisk2020experience, Jelinek2023language}.
For example, an agent could make a cup of coffee following a series of primitive instructions including detailed descriptions of the target object from a Large Language Model (LLM) planner like GPT-4~\cite{openai2023gpt4}.  
Within the process, visual language grounding plays a key role in linking each step instruction with the physically observed object~\cite {ahn2022can, song2022human}.
Therefore, visual language grounding with 3D objects is the indispensable means of enabling an agent to interact with the real world. 
A limited scale of high-quality vision-language paired data hinders the development of visual language grounding technology, especially 3D vision language grounding.
To address this problem, existing work attempts~\cite{thomason2022language,song2023scenedriven,corona2022voxel,mitra2023comparative} to use multi-view perception or external priors, which need extra data cost and existing domain gap caused by pre-trained feature encoders in fixed settings. 
In this paper, we give an exploration from a domain adaptation perspective for the language grounding task inspired by domain adaptation in parameter-efficient tuning of large language model~\cite{tai2020exbert, diao2021taming, vstefanik2022adaptor,malik2023udapter,diao2023mixture, gururangan2020don,hu2021lora}. 

\begin{figure}[t]
    \vspace{-10pt}
    \centering
    \includegraphics[width=4.6in]{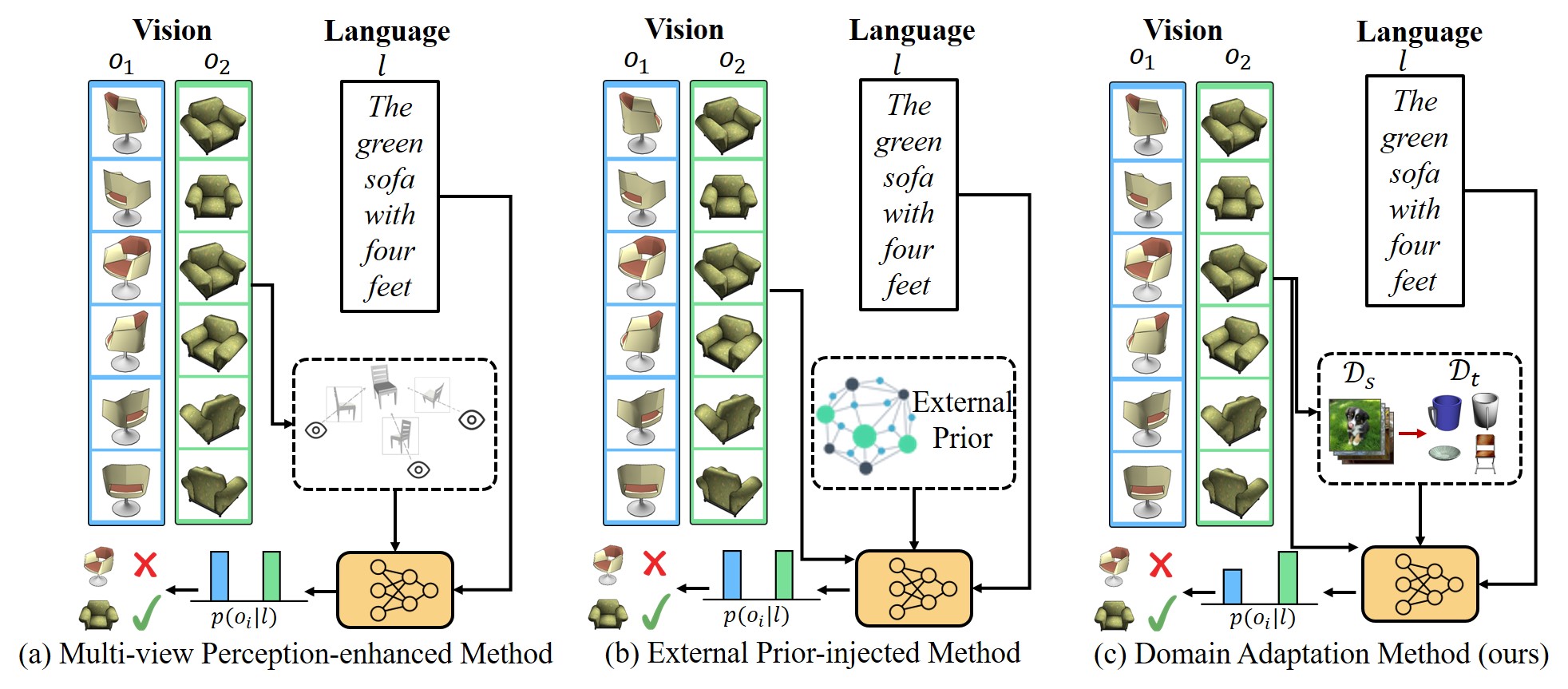}
    \caption{The comparison between existing works and our model. 
    Existing works focus on (a) multi-view perception and (b) external prior.
    (c) We approach language grounding from domain adaptation.
}
    \label{fig:pipeline_overview}
    \vspace{-15pt}
\end{figure}

As shown in Fig.~\ref{fig:pipeline_overview}, existing research in language grounding mainly focuses on two lines including the multi-view perception-enhanced method (Fig.~\ref{fig:pipeline_overview} (a)) and the external prior-injected method (Fig.~\ref{fig:pipeline_overview} (b)). 
For the former, Thomason et al.~\cite{thomason2022language} and Mitra et al.~\cite{mitra2023comparative} propose a view-based method to improve the prediction accuracy. 
Thomason et al.~\cite{thomason2022language} design an auxiliary task of viewpoint angle estimation to enhance 3D object understanding. 
Mitra et al.~\cite{mitra2023comparative} design a multi-view transformer to fuse visual features and textual features in shared space. 
For the latter, Corona et al.~\cite{corona2022voxel} propose a voxel-informed method by using pre-trained 3D volumetric generative model LegoFormer based on the view of 3D objects~\cite{yagubbayli2021legoformer}. 
Song et al.~\cite{song2023scenedriven} construct explicit scene-driven multimodal knowledge graph ManipMob-MMKG to design a knowledge-enhanced method. 
In all, current methods still depend heavily on viewpoints or external priors.


In the representation learning aspect, existing works usually adopt vision-language feature encoders using freezing parameter mode pre-trained on a source domain, which cannot work well for 3D language grounding tasks in multimodal alignment due to domain gap.
Based on these findings, we propose a novel multimodal domain adaptation method \textbf{D}omain \textbf{A}daptation for \textbf{L}anguage \textbf{G}rounding named \textbf{DA4LG}, to improve 3D object-level understanding and multimodal alignment, which does not require extra visual or textual data, as shown in Fig.~\ref{fig:pipeline_overview}(c).
Given the language similarity between the source domain (e.g., the WebImageText domain, where CLIP is pre-trained~\cite{radford2021learning}) and the target domain (e.g., language grounding domain), coupled with the generalization capabilities of pre-trained language models~\cite{wang2023improving,devillers2021does,hao2019visualizing}, DA4LG concentrates on domain adaptation within the vision feature.
Specifically, we design the pseudo-siamese visual encoder network~\cite{gong2023cross} to realize domain adaptation, in which one visual encoder subnetwork is used to learn domain-specific 3D visual representation, named \textbf{Domain-specific Encoder}, while another one is freezing to encode visual representations associated with the source domain.
\syx{For training the model, we design two auxiliary tasks with the main language grounding task (\textbf{LGR}) to learn cross-modal representation. 
The first task is to distinguish different objects using vision and language contrastive learning, while the second involves regenerating the input text from multi-modal fused features.
We evaluate DA4LG via the language grounding dataset SNARE proposed by Thomason et al.\cite{thomason2022language}, which discriminates natural language descriptions with 3D ShapeNet~\cite{chang2015shapenet} objects.
DA4LG achieves state-of-the-art (SOTA) performance in both single-view and multi-view settings.}
Additionally, through simulation experiments, DA4LG demonstrates generalization and robustness compared to existing models.
Compared to multi-view perception-enhanced methods such as Thomason et al.~\cite{thomason2022language} and Mitral et al.~\cite{mitra2023comparative}, our DA4LG is immune to the number of viewpoints or the selection of viewpoints. 
Compared to external prior-injected methods such as Coronal et al.~\cite{corona2022voxel} and Song et al.~\cite{song2023scenedriven}, our proposed method requires only adapting the visual encoder as a cloned module with limited parameter training, without the need for external prior injection. 
This approach presents a clear advantage in terms of reducing the model's parameter size while simultaneously enhancing its reliability.

Our main contributions can be summarized as follows:

\begin{enumerate}
\syx{    
\item We introduce a novel domain adaptation method (DA4LG) using multi-task learning to reduce visual domain gap in vision-language aligned representation for language grounding with 3D objects.

\item DA4LG demonstrates \textbf{SOTA} performance, achieving $83.8 \%$ accuracy in a single-view setting and $86.8 \%$ accuracy in a multi-view setting on the language grounding benchmark SNARE~\cite{thomason2022language}.

\item We conduct the simulation 3D object grounding experiment including a grounding environment and a test set extended from Lang-SHAPE~\cite{song2023learning} referred to as \textbf{Simulation-SNARE}. 
The results indicate that DA4LG has an obvious advantage in the robustness and generalization of applications compared to existing models.}
\end{enumerate}

\begin{figure}[!t]
\centering
\includegraphics[width=0.76\linewidth]{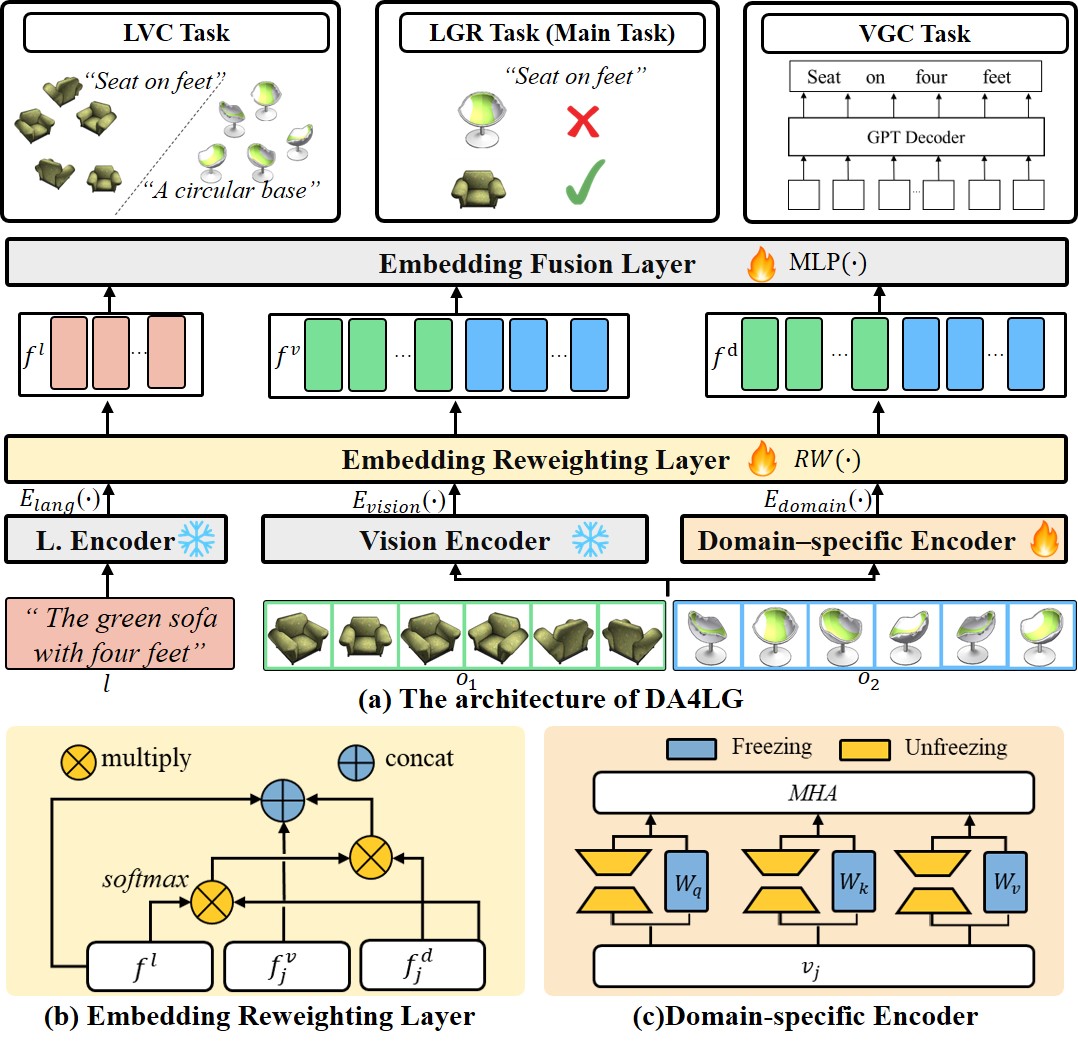}
\caption{The framework of DA4LG. DA4LG is comprised of Encoder Layer, Embedding Reweighting Layer and Embedding Fusion Layer.
Encoder Layer contains three encoders: Language Encoder (L. Encoder), Vision Encoder, and Domain-specific Encoder. 
The snowflake and fire denote the freezing and unfreezing respectively. }
\label{fig:model_overview}
\end{figure}

\section{Related Work}

\noindent\textbf{Language Grounding}.
Language grounding with 3D can be classified into two categories: language grounding for navigation and language grounding for interaction.
Language grounding for navigation focuses on training an agent to follow a set of natural language instructions to navigate towards a target object in an environment, as discussed by Chen et al~\cite{chen2022language}. 
This task is relevant to scene understanding including object localization~\cite{achlioptas2020referit3d,roh2022languagerefer,zhang2023multi3drefer,huang2022multi}, visual language navigation~\cite{shrivastava2022visitron,schumann2022analyzing,miyanishi2023cross3dvg,guo2023viewrefer}.

Recently, research has emerged in the language grounding for interaction with 3D objects.
ShapeGlot~\cite{achlioptas2019shapeglot} investigates how linguistic expressions capture detailed differences in the shapes of common objects based on their 3D representations. 
Akula et al.~\cite{akula2021mind} propose the use of neural module networks that share weights and exploit associations between similar textual contexts, such as "\textit{dark cube on the left}" and "\textit{black cube on the left}". 
SNARE~\cite{thomason2022language} presents a challenge by requiring the identification of referent objects that are highly similar to distractors. 
This task encompasses various perspectives of objects in 3D space from ShapeNet~\cite{chang2015shapenet}, thereby increasing the complexity of the problem in language referencing.

Current research encompasses two main approaches: the multi-view perception-enhanced method and the external prior-injected method.
The former employs a multi-view framework with 3D objects to enhance prediction accuracy such as MAGiC~\cite{mitra2023comparative}, LAGOR~\cite{thomason2022language}, among others.
The latter approach integrates external priors to augment task performance, including  VLG~\cite{corona2022voxel}, LOCKET~\cite{song2023scenedriven}, among others.

\noindent\textbf{Domain Adaptation}.
Recent research has placed particular emphasis on adaptors designed for domain adaptation in the natural language processing (NLP). 
Hu et al.~\cite{hu2021lora} propose Low-Rank Adaptation(LoRA), a novel approach involving freezing pre-trained model weights and introducing trainable rank decomposition matrices as adapter into each layer of the Transformer architecture. 
Effective adaptation strategies based on this are introduced to bridge the domain gap in NLP tasks~\cite{tai2020exbert, diao2021taming, vstefanik2022adaptor,malik2023udapter,diao2023mixture, gururangan2020don,hu2021lora}.
Inspired by the existing research, we incorporate domain adapter into multimodal tasks and explore its utility in adapting language grounding with 3D objects, extending the application of it beyond traditional NLP tasks.

\section{Proposed Method}

\noindent\textbf{Domain Adaptation.}
\label{optimization_objective}
Given a pre-trained network model $\mathcal{F}_{s}(\cdot)$ on a source task $\mathcal{T}_{s}$ (e.g., CLIP pre-trained model~\cite{radford2021learning}) in the source domain $\mathcal{D}_{s}$ (e.g., the WebImageText domain where CLIP is pre-trained on~\cite{radford2021learning}), and a set of training examples with associated labels in the target domain $\mathcal{D}_{t}$ for target task $\mathcal{T}_{t}$  (i.e. the language grounding task).
Our goal is to create an adaption strategy to promote the performance of target predictive function $\mathcal{F}_{t}(\cdot)$ in $\mathcal{D}_{t}$ for $\mathcal{T}_{t}$ through leveraging the $\mathcal{F}_{s}(\cdot)$ in $\mathcal{D}_{s}$ and $\mathcal{T}_{s}$~\cite{csurka2017domain}.

\noindent\textbf{Task Definition.} 
In the language grounding task with a 3D object, given a language description $l$ as input, our objective is to identify the best matching object $\hat{o}$ in a set of candidate objects $\mathcal{O}=\{o_i\mid i \in \{1,2\}\}$~\cite{thomason2022language}:
\begin{equation}
\small
\label{eq:definition}
\scalebox{1.0}{$
\begin{aligned}
 \hat{o} = \mathop{\arg \max }\limits_{{o_i} \in \mathcal{O}} \; p(o_i | l),
\end{aligned}
$}
\end{equation}
where $p(o_i | l)$ denotes the conditional probability of the target object given the language description.
We expect our output $\hat{o}$ to closely align with the ground truth $o^{*}$ as much as possible.

\subsection{Domain Adaptation for Language Grounding}
\label{sec:domain_aware_representation}
\noindent \textbf{Overall of Network Structure.}\quad
As shown in Fig.~\ref{fig:model_overview} (a), 
DA4LG comprises three encoders: a Vision Encoder ($E_{vision}(\cdot)$), a Language Encoder ($E_{lang}(\cdot)$), and a Domain-specific Encoder ($E_{domain}(\cdot)$). 
Additionally, it includes an Embedding Reweighting Layer ($RW(\cdot)$) and an Embedding Fusion Layer ($MLP(\cdot)$).
Therefore Eq.~\ref{eq:definition} can be rewritten as:
\begin{equation}
\scalebox{1.0}{$
\begin{aligned}
 f^v_j &= E_{vision}(v_j), \\
 f^d_j &= E_{domain}(v_j), \\
 f^l &= E_{lang}(l), \\
 Score &= MLP(RW(f^d_j), f^v_j, f^l), \\
 \hat{o} &= \mathop{\arg \max } \limits_{ {o_i} \in \mathcal{O} } \; Score_{o_i},
\end{aligned}
$}
\end{equation}
where the object $o$ is projected into a set of images $\mathcal{V} = \{v_1,..., v_j\}$ from discrete viewpoints.
$E_{vision}$ and $E_{lang}$ are based on the pre-trained encoder in the freezing state.
We compute the $Score_{o_i}$ to determine the output label $\hat{o}$.


\noindent \textbf{Domain-specific Encoder.}\quad 
To reduce the domain gap between $\mathcal{D}_{s}$ and $\mathcal{D}_{t}$, we design a Domain Vision Transformer named \textbf{Domain-specific Encoder} to encode the $\mathcal{V}$, as shown in Fig.~\ref{fig:model_overview} (c).
Domain-specific Encoder is pre-trained on $\mathcal{D}_{s}$ (e.g., the WebImageText domain~\cite{radford2021learning}).
Compared to vanilla Vision Transformer~\cite{dosovitskiy2020image}, we incorporate low-rank matrices $W^{A}_{*}, W^{B}_{*}$ as domain adapters in the $W_q$, $W_k$, $W_v$ of multi-head attention layer $MHA$.
Different from the adapters employed for parameter-efficient tuning in NLP, the domain adapters in DA4LG are designed to capture the domain-specific representation.
In Domain-specific Encoder, all other parameters are freezing except the $W^{A}_{*}, W^{B}_{*}$  while training.
${v_j}$ is input into the $MHA$ to build domain features $f_j^d$, 
\begin{equation}
\small
\scalebox{1.0}{$
\begin{aligned} 
Q  &=  W_q \cdot v_j + W^{A}_q  \cdot  W^{B}_q  \cdot v_j ,\\
K  &= W_k  \cdot v_j + W^{A}_k  \cdot  W^{B}_k \cdot  v_j ,\\
V  &= W_v  \cdot v_j  + W^{A}_v \cdot   W^{B}_v \cdot  v_j ,\\
f_j^d &= MHA(Q, K, V).
\end{aligned}
$}
\end{equation}

\noindent \textbf{Embedding Reweighting Layer.}\quad
As shown in Fig.~\ref{fig:model_overview} (b), we use $RW(\cdot)$ to adjust the $f_j^d$ and reduce the impact of those features that are irrelated to $f^l$.
Specifically, given the $f_j^d$ corresponding to the $j$-th view $v_{j}$ and the description $f^l$, we compute the cosine similarity with $sim(f_j^d, f^l)$ and get the weighted combination of domain features $f_j^d = f_j^d \cdot softmax(sim(f_j^d, f^l))$.

\noindent \textbf{Embedding Fusion Layer.}\quad
To enhance multimodal alignment and construct joint features, we employ the aggregate operation $agg$ in Embedding Fusion Layer to build the vision features $f^v = agg(f_j^v)$ and domain features $f^d = agg(f_j^d)$, where $agg$ is max-pooling.
We concatenate the features $f = [ f^l, f^v, f^d ]$ which is fed into a multi-layer perceptron $MLP(\cdot)$ to compute $Score$.


\subsection{Multi-task Learning}


As shown in Fig.~\ref{fig:model_overview} (a), the DA4LG framework incorporates three different tasks: the Language Grounding~(LGR) Task, the Vision-Language Contrastive~(VLC) Task, and the Vision Grounding Caption~(VGC) Task in multi-task learning.
The LGR Task is designed as the primary task following the existing works~\cite{thomason2022language}.
The VLC and VGC tasks serve as auxiliary tasks to optimize the training objectives inspired by BLIP-2~\cite{li2023blip2}.

\noindent \textbf{Language Grounding Task (LGR Task).}\quad
The primary task is the LGR task, which involves predicting the target. 
We feed $Score$ to predict the ground truth label $o^{*}$ and apply the binary cross-entropy loss for optimization:
\begin{equation}
\small
\scalebox{1.0}{$
\begin{aligned} 
\mathcal{L}_{LGR} =  -\mathbb{E} [o^{*}log(Score) + (1 - o^{*})log(1 - Score)].
\end{aligned}
$}
\end{equation}

\noindent \textbf{Vision-Language Contrastive Task (VLC Task).}\quad
We propose the VLC task to learn an embedding that distinguishes samples from two different distributions.
At each step, we sample some positive or negative pairs $(l, o)$ during training.
Specifically, samples from matched pairs are termed positives, whereas those from unmatched pairs are termed negatives.
We use cosine similarity, denoted as $s({f^l, f^o})$, to measure the alignment between the language features $f^l$ and the object features $f^o$, where $f^o = agg([f_j^v, f_j^d])$.
We optimize this function to correctly select a single positive description sample $l$ with $\Phi$ negative object samples and calculate the contrastive loss for object description:
\begin{equation}
\small
\scalebox{1.0}{$
\begin{aligned} 
\mathcal{L}_{con}^{o\to l } = -\mathbb{E} [\text{log}\frac{s(f^l,f^{o_p})}{s(f^l,f^{o_p}) + \sum_{\varphi=1}^{\Phi}s(f^l,f^{o_n^\varphi})}  ], 
\end{aligned}
$}
\end{equation}
where $f^{o_p}$, $f^{o_n^\varphi}$ are the features of the positive and  negatives sample for 3D objects.
Similarly, we can obtain the contrastive loss for the description-object pairs:
\begin{equation}
\small
\scalebox{1.0}{$
\begin{aligned} 
\mathcal{L}_{con}^{l\to o } = -\mathbb{E} [\text{log}\frac{s_j(f^{l_p}, f^o)}{s(f^{l_p}, f^o) + \sum_{\varphi=1}^{\Phi}s(f^{l_n^\varphi}, f^{o})}  ].
\end{aligned}
$}
\end{equation}
And the VLC loss is denoted as $\mathcal{L}_{VLC} = \mathcal{L}_{con}^{l\to o } + \mathcal{L}_{con}^{o\to l }.$

\noindent \textbf{Vision Grounding Caption Task (VGC Task).}\quad
Given the $f^o$ and $l$, we design the VGC task to generate textual descriptions based on the freezing GPT-$2$~\cite{radford2019language}.
Here, $l$ represents a sequence of tokens denoted as $\{c_i, i = [1, ..., N]\}$.
Our training objective is to predict the caption tokens conditioned on the output token in auto-regressive. 
The training loss for the VGC task is formulated as:
\begin{equation}
\small
\scalebox{1.0}{$
\begin{aligned} 
\mathcal{L}_{VGC} =-\mathbb{E} [\sum_{i=1}^{N} \text{log} \; p(c_i| f^o,c_1, ..., c_{i-1}) ].
\end{aligned}
$}
\end{equation}
During the multi-task learning phrase, our target loss function is
\begin{equation}
\small
\scalebox{1.0}{$
\begin{aligned} 
\mathcal{L} =  \mathcal{L}_{LGR} + \mathcal{L}_{VLC} +   \mathcal{L}_{VGC}.
\end{aligned}
$}
\end{equation}

\begin{table*}[t]
\caption{Performance of our model and existing work on the SNARE dataset. 
Absence of parentheses signifies that the original paper does not report standard deviations.
Standard deviations over five seeds are shown in parentheses (the same as below).
}
    \label{tab:main_comparsion}
    \centering
    \scalebox{0.80}{
    \begin{tabular}{@{}c|c|ccc|ccc@{}}
\toprule
\multirow{2}{*}{Model}                                  & \multirow{2}{*}{Method} & \multicolumn{3}{c|}{Validation}             & \multicolumn{3}{c}{Test}             \\ \cmidrule(l){3-8} 
                                                        &                         & Visual     & Blind      & All        & Visual     & Blind      & All        \\ \midrule
ViLBERT~\cite{lu2019vilbert}      & External Prior          & 89.5       & 76.6       & 83.1       & 80.2       & 73         & 76.6       \\
CLIP~\cite{radford2021learning}   & External Prior          & 83.7       & 65.2       & 74.5       & 80.0       & 61.4       & 70.9       \\
MATCH~\cite{thomason2022language} & External Prior          & 89.2 (0.9) & 75.2 (0.7) & 82.2 (0.4) & 83.9 (0.5) & 68.7 (0.9) & 76.5 (0.5) \\
LAGOR~\cite{thomason2022language} & Multi-view Perception   & 89.8 (0.4) & 75.3 (0.7) & 82.6 (0.4) & 84.3 (0.4) & 69.4 (0.5) & 77.0 (0.5) \\
VLG~\cite{corona2022voxel}        & External Prior          & 91.2 (0.4) & 78.4 (0.7) & 84.9 (0.3) & 86.0       & 71.7       & 79.0       \\
BLIP2~\cite{li2023blip2}          & External Prior          & 51.2       & 50.9       & 51.5       & -          & -          & -          \\
LOCKET~\cite{song2023scenedriven} & External Prior          & 90.9       & 78.4       & 84.7       & 86.1       & 71.5       & 79.0       \\
MAGiC~\cite{mitra2023comparative} & Multi-view Perception   & 92.1 (0.4) & 81.3 (0.9) & \textbf{86.8 (0.5)} & 87.7       & 75.4       & 81.7       \\ \midrule
\textbf{DA4LG}(ours)                   & Domain Adaptation       & 91.8 (0.3) & \textbf{81.8 (0.6)} & \textbf{86.8 (0.5)} & \textbf{88.5}          & 75.0          & \textbf{81.9}          \\ \bottomrule
\end{tabular}
}
\vspace{-10pt}
\end{table*}

\section{Experiment Design}
To evaluate our proposed method, we investigate four key research questions (\textbf{RQ}s).
The \textbf{RQ1} concerns the DA4LG superior to the baselines.
The \textbf{RQ2} and \textbf{RQ3} focus on how to adopt DA4LG efficiently.
The \textbf{RQ4} concerns the generalization of DA4LG in \syx{the downstream} simulation environment.

\begin{enumerate}
    \item[] $\bullet$ \syx{\noindent \textbf{RQ1}: What are the advantages of our DA4LG compared to other methods for the language grounding task?}

\item[] $\bullet$ \syx{\noindent \textbf{RQ2}: Which training policy in Domain-specific Encoder can obtain the better performance with efficient parameters?}

\item[] $\bullet$ \noindent \textbf{RQ3}: How do different learning tasks in DA4LG affect the language grounding performance?

\item[] $\bullet$ \syx{\noindent \textbf{RQ4}: Can our DA4LG perform more effectively deployed in a downstream task by a simulation environment compared to other methods?} 
\end{enumerate}

\subsection{Baseline Models}
We conduct a comparative analysis between DA4LG and various public baselines, as summarized in Table~\ref{tab:main_comparsion}. 
Two primary approaches are utilized in current research: the multi-view perception-enhanced method, exemplified by MAGiC~\cite{mitra2023comparative} and LAGOR~\cite{thomason2022language}, and the external prior-injected method, represented by ViLBERT~\cite{lu2019vilbert}, MATCH~\cite{thomason2022language}, VLG~\cite{corona2022voxel}, CLIP~\cite{radford2021learning}, LOCKET~\cite{song2023scenedriven},  and BLIP2~\cite{li2023blip2}. We list these baselines below:

\begin{itemize}
\item[] $\bullet$ \textbf{LAGOR} adopts a multi-task learning approach by predicting the canonical viewing angle for individual view images.

\item[] $\bullet$ \textbf{MAGiC} performs joint reasoning over candidate referent objects, considering each object from multiple possible perspectives.

\item[] $\bullet$ \textbf{MATCH} and \textbf{ViLBERT} uses CLIP-ViT and ViLBERT respectively to encode the views of each object. 
An MLP is trained to assign scores based on the encoded views and language description embedding

\item[] $\bullet$ \textbf{CLIP} uses the cosine distance in CLIP embedding between visual and language features to pick the object in the lowest distance.

\item[] $\bullet$ \textbf{BLIP-2} is the multimodal LLM-based method in zero-shot setting.

\item[] $\bullet$ \textbf{VLG} leverages implicit 3D prior information from predicted volumetric voxel maps to improve language grounding performance by LegoFormer~\cite{yagubbayli2021legoformer}. 

\item[] $\bullet$ \textbf{LOCKET} is a knowledge enhancement method that employs a graph convolutional network to encode a multimodal knowledge graph. 

\end{itemize}

\subsection{Implementation Details}

\subsubsection{Training and Inference Details.}
We employ the Adam optimizer with weight decay $5e^{-4}$.
The batchsize is $64$, training epoch is $60$, and learning rate is $5e^{-3}$.
Experiments are implemented with CUDA $11.2$ and PyTorch $1.7.1$ and run on one NVIDIA RTX4090.
In DA4LG, we employ the vision and language encoders from CLIP ViT-B/32~\cite{radford2021learning} as the Vision Encoder and Language Encoder, respectively. 
Domain-specific Encoder is initialized from the vision encoder in CLIP ViT-B/32.

\subsubsection{Benchmark Datasets.} 
We train and evaluate our proposed method on SNARE dataset~\cite{thomason2022language} which is split into training, validation, and test sets following existing works~\cite{thomason2022language}. 
SNARE is a benchmark for choosing the correct object with small differences in multiple views given a language description.
Each data in the set has the label of visual or blind.
The visual label means a comprehensive understanding of the object, providing relevant visual cues to guide the grounding process (e.g., “\textit{classic armchair with white seat}”).
The blind labels predominantly focus on the object's shape and specific distinguishing attributes, intentionally omitting color and other visual characteristics (e.g., “\textit{oval back and vertical legs}”). 
The training set consists of $207$ categories, $6,153$ objects, and $39,104$ references. 
The validation set contains $7$ categories, $371$ objects, and $2,304$ references. 
The test set consists of $48$ categories, $1,357$ objects, and $8,751$ references.

\subsubsection{Simulation Details.}
We build a new simulation benchmark by sampling the objects from the existing 3D datasets Lang-SHAPE~\cite{song2023learning} and annotating them with the original data annotation process utilized in SNARE, which we refer to as Simulation-SNARE. 
Specifically, the objects and domains in Lang-SHAPE are identical to those in SNARE, both of which are derived from ShapeNet~\cite{chang2015shapenet}. 
We follow the original data annotation process utilized in SNARE. 
Simulation-SNARE consists of $327$ objects and $2,876$ references, where $634$ classified as visual and $2,242$ as blind, following the configuration of SNARE.
We deploy the Simulation-SNARE into the simulation environment MuJoCo~\cite{todorov2012mujoco}. 
We replicate the existing methods CLIP, ViLBERT, MATCH, VLG, and LAGOR with open-source code, deploying these methods along with DA4LG in the simulation world.

\subsubsection{Metric.}
The metric for this task is the accuracy ($\%$) of the predictions in correctly identifying the object referred to by the language description from two candidate objects. 
We calculate the accuracy across all sets, the visual subset and the blind subset respectively.
Additionally, we assess the model's parameter efficiency by calculating its parameter size.

\section{Result Analysis}
\subsection{Benchmark Comparisons}
\label{results_analysis}

\subsubsection{Comparisons with Existing methods.}
\label{advantages_of_da4lg}

\begin{table}[t]
\caption{Performance on the model with single-view. 
*We obtain the reported results from the curve~\cite{mitra2023comparative}, which is below but close to $82.0 \%$. 
**The results come from our reproduction.
}
    \label{tab:single_view_study}
    \centering
    \scalebox{0.80}{
   \begin{tabular}{@{}c|c|ccc@{}}
\toprule
\multirow{2}{*}{Models}                                          & \multirow{2}{*}{Method} & \multicolumn{3}{c}{Validation}                                 \\ \cmidrule(l){3-5} 
                                                                 &                         & Visual     & Blind      & All                                  \\ \midrule
CLIP~\cite{radford2021learning}            & External Prior          & 79.0       & 63.0       & 71.1                                 \\
MATCH~\cite{thomason2022language}          & External Prior          & 88.4 (0.4) & 73.3 (0.6) & 80.9 (0.4)                           \\
VLG~\cite{corona2022voxel}$^{\star \star}$ & External Prior          & 89.3 (0.7) & 74.1 (0.9) & 81.8 (0.6)                           \\
MAGiC~\cite{mitra2023comparative}          & Multi-view Perception   & -          & -          & 82.0$^{\star}$                       \\
\textbf{DA4LG (ours)}                           & Domain Adaptation       & \textbf{90.1 (0.5)} & \textbf{77.1 (0.8)} & \textbf{83.8 (0.5)} \\ \bottomrule
\end{tabular}
}
\vspace{-10pt}
\end{table}

To answer \textbf{RQ1}, we conduct a comparative analysis in multi-view and single-view settings and our proposed model DA4LG outperforms in all settings compared to baselines.
In the \textbf{multi-view setting}, Table~\ref{tab:main_comparsion} demonstrates that DA4LG achieves the best improvements in validation performance compared to the \textbf{External Prior} methods utilizing the same backbone CLIP ViT-B/32. 
Specifically, DA4LG exhibits enhancements of $12.3 \%$ ($74.5 \% \rightarrow 86.8 \%$), $4.6 \%$ ($82.2 \% \rightarrow 86.8 \%$), $2.1 \%$  ($84.7 \% \rightarrow 86.8 \%$), and $1.9 \%$ ($84.9 \% \rightarrow 86.8 \%$) compared to CLIP, MATCH, LOCKET, and VLG methods, respectively. 
Specifically, the VLG model requires $168$ million training parameters, while our method requires only less than half of the VLG model's parameters ($79.5$ million). 
This demonstrates the superior parameter efficiency of DA4LG.
Contrary to the LOCKET model, which demands a knowledge graph for extensive data, our method eliminates the necessity for such a structure.
When employing ViLBERT as the backbone, our model achieves a $3.7 \% (83.1\% \rightarrow 86.8\%)$ improvement in validation accuracy. 
Furthermore, when utilizing a multimodal LLM in a zero-shot approach, our model exhibits an increase in performance exceeding $30 \%$.
The \textbf{Multi-view Perception} method includes LAGOR and MAGiC. 
DA4LG demonstrates improvements, with a $4.2 \% (82.6 \% \rightarrow 86.8 \%)$ improvement in validation compared to LAGOR.
DA4LG achieves a validation score of $86.8 \%$ on the validation set and $81.9 \%$ on the test set, tying with the existing SOTA model MAGiC in the validation and surpassing it by $0.2 \%$ in the test.

The performance of models in the \textbf{single-view setting} are shown in Table~\ref{tab:single_view_study}.
DA4LG achieves the best result among all methods, surpassing the existing SOTA model MAGiC by over $1.8 \% (82.0 \% \rightarrow 83.8 \%)$ in validation.
DA4LG in the single-view setting achieves better performance than several models in the multi-view setting. 
Specifically, DA4LG exhibits improvements of $9.3 \% (74.5 \% \rightarrow 83.8 \%)$, $1.6 \% (82.2 \% \rightarrow 83.8 \%)$, $1.2 \% (82.6 \% \rightarrow 83.8 \%)$, and $0.7 \% (83.1 \% \rightarrow 83.8 \%)$ over the CLIP, MATCH, LAGOR, and ViLBERT models in the multi-view setting respectively.
This study shows an aligned multimodal feature in the single-view setting can enhance the overall performance of the model in the language grounding task.
We believe that the observed improvement in performance can be attributed to the multimodal alignment within the target domain. 
In the following subsection, we provide ablation studies and visualizations to further demonstrate this result.

\subsubsection{Training Policy in Domain-specific Encoder.}

\begin{table}[t]
\caption{Performance and training parameters across training mode in Domain-specific Encoder. 
M means one million and \#Param means the training parameter size (the same as below). 
}
    \label{tab:domain_adaptation_strategy}
    \centering
    \scalebox{0.80}{
   \begin{tabular}{@{}c|ccc|c@{}}
\toprule
\multirow{2}{*}{Strategies}    & \multicolumn{3}{c|}{Validation}                                 & \multirow{2}{*}{
\begin{tabular}[c]{@{}c@{}}\#Param \end{tabular}} \\ \cmidrule(lr){2-4}
                              & Visual              & Blind               & All                 &                                                                                      \\ \midrule

Freezing-Param            & 89.9 (0.5)          & 79.8 (0.4)          & 84.9 (0.4)          & 79.2M                                                                                \\
Full-Param                  & 86.4 (0.7)                            & 75.0 (0.2)                            & 80.8 (0.6)         & 254M                                                                                 \\
Partial-Param & 89.9 (0.4)           & 77.7 (0.6)           & 83.8 (0.5)           & 107M                                                                                 \\
Domain-Adapter                & \textbf{91.8 (0.3)} & \textbf{81.8 (0.6)} & \textbf{86.8 (0.5)} & 79.5M                                                                                \\ \bottomrule
\end{tabular}
}
\vspace{-10pt}
\end{table}

To answer \textbf{RQ2}, we explore the different training mode and the source domains in Domain-specific Encoder.
For training modes, we conduct a comparative analysis of following four scenarios where the Domain-specific Encoder is initialized using the same pre-trained parameters:

\begin{itemize}
\item[] $\bullet$  \textbf{Freezing-Param}:  The parameters of Domain-specific Encoder is fixed.

\item[] $\bullet$ \textbf{Full-Param}: The full parameter in the Domain-specific Encoder is trainable.

\item[] $\bullet$  \textbf{Partial-Param}: Inspired by Sun et al.~\cite{sun2022unfreeze}, only the last two layers of the Domain-specific Encoder are updated during the training stage.

\item[] $\bullet$ \textbf{Domain-Adapter}: The Domain-specific Encoder is training with domain adaptor mentioned in Section~\ref{sec:domain_aware_representation} following Low-Rank Adaptation (LoRA)~\cite{hu2021lora}.
\end{itemize}

As presented in Table~\ref{tab:domain_adaptation_strategy}, Domain-Adapter achieves the best performance compared to other training modes, with an accuracy of $86.8 \%$ on the validation set with the limited training parameters.
The Full-Param and Partial-Param strategies achieved performance of $80.8 \%$ and $83.8 \%$ in the validation set respectively. 
However, they come with the drawback of increased training parameters compared to the Freezing-Param and Domain-Adapter methods. 


\begin{table*}[t]
\caption{Comparative performance of Domain-Specific Encoders initialized with models from various source domains.
}
    \label{tab:domain_knowledge_ablation}
    \centering
    \scalebox{0.80}{
  \begin{tabular}{@{}c|c|c|ccc@{}}
\toprule
Models    & Source Domain                                                 & \#Param & Visual     & Blind      & All        \\ \midrule
Scratch-B & -                                                             & 87M     & 86.4 (0.7) & 76.7 (0.2) & 81.6 (0.6) \\
BLIP-B    & BLIP~\cite{li2022blip}                  & 86M     & +2.8 (0.4) & +4.8 (0.5) & +3.0 (0.6) \\
ViT-B     & ImageNet21K~\cite{deng2009imagenet}     & 87M     & +3.4 (0.5) & +4.8 (0.7) & +4.0 (0.4) \\
CLIP-B    & WebImageText~\cite{radford2021learning} & 87M     & +5.4 (0.3) & +5.1 (0.6) & +5.2 (0.5) \\ \midrule
Scratch-L & -                                                             & 303M    & 86.9 (0.5) & 74.2 (0.8) & 80.6 (0.5) \\
BLIP-L    & BLIP~\cite{li2022blip}                   & 303M    & +3.0 (0.8) & +3.9 (0.6) & +3.3 (0.5) \\
ViT-L     & ImageNet21K~\cite{deng2009imagenet}      & 304M    & +4.3 (0.6) & +3.8 (0.8) & +4.1 (0.5) \\
CLIP-L    & WebImageText~\cite{radford2021learning} & 303M    & +3.8 (0.6) & +5.1 (0.7) & +4.4 (0.4) \\ \bottomrule
\end{tabular}
}
\vspace{-10pt}
\end{table*}

We select four source domains to initialize Domain-specific Encoder for \textbf{RQ2}.
Table~\ref{tab:domain_knowledge_ablation} demonstrates that a Domain-specific Encoder pre-trained on source domains exhibits better performance compared to training from scratch.
Additionally, employing a larger-parameter Vision Transformer as the domain-specific encoder does not significantly improve the performance of the DA4LG.
Domain-specific Encoder using CLIP-B (pre-trained on a specific domain WebImageText) demonstrates a significant improvement from $81.6 \%$ to $86.8 \%$ compared to Scratch-B. 
The performance of ViT-B and BLIP-B, which pre-trained on ImageNet domain and BLIP domain, demonstrates an enhancement by $3.0 \%$ and $4.0 \%$, respectively.
The performance for CLIP-L and Scratch-L are $85.0 \%$ and $80.6 \%$, respectively. 
These figures do not show a significant improvement compared to the metrics for CLIP-B and Scratch-B, which are $86.8 \%$ and $81.6 \%$, respectively. 
Therefore, initializing with CLIP-B is the preferred choice for domain-specific applications.

\subsubsection{Ablation Study.}

\begin{table}[t]
\caption{The ablation study of multi-task learning and encoders. 
The terms Language, Vision, and Domain refer to the Language Encoder, Vision Encoder, and Domain-specific Encoder, respectively.
}
    \label{tab:ablation_study}
    \centering
    \scalebox{0.80}{
   \begin{tabular}{@{}ccc|ccc|ccc@{}}
\toprule
\multicolumn{3}{c|}{Input Encoders}                                                         & \multicolumn{3}{c|}{Task}                                                             & \multicolumn{3}{c}{Validation}                                  \\ \midrule
Language                      & Vision                      & Domain                      & LGR                       & VLC                         & VGC                         & Visual              & Blind               & All                 \\ \midrule
$\checkmark$ & $\checkmark$   & $\checkmark$   & $\checkmark$ & $\times$ & $\times$ & 89.8 (1.1)          & 74.0 (0.4)          & 81.9 (0.8)          \\
$\checkmark$ & $\checkmark$   & $\checkmark$   & $\checkmark$ & $\checkmark$   & $\times$ & 91.0 (0.5)          & 79.8 (0.4)          & 85.4 (0.1)          \\
$\checkmark$ & $\checkmark$   & $\checkmark$   & $\checkmark$ & $\times$ & $\checkmark$   & 90.9 (0.4)          & 78.9 (0.7)          & 85.0 (0.5)          \\ \midrule
$\checkmark$ & $\checkmark$   & $\times$ & $\checkmark$ & $\checkmark$   & $\checkmark$   & 89.3 (0.8)          & 76.3 (0.6)          & 82.9 (0.7)          \\
$\checkmark$ & $\times$ & $\checkmark$   & $\checkmark$ & $\checkmark$   & $\checkmark$   & 89.0 (0.6)          & 73.9 (0.7)          & 81.5 (0.6)          \\ \midrule
$\checkmark$ & $\checkmark$   & $\checkmark$   & $\checkmark$ & $\checkmark$   & $\checkmark$   & \textbf{91.8 (0.3)} & \textbf{81.8 (0.6)} & \textbf{86.8 (0.5)} \\ \bottomrule
\end{tabular}
}
\vspace{-10pt}
\end{table}

For \textbf{RQ3}, we perform an ablation study analysis to investigate the influence of different tasks and encoders in DA4LG.
For different tasks, Table~\ref{tab:ablation_study} demonstrates the effectiveness of different tasks and encoders within DA4LG. 
The DA4LG with all tasks and encoders achieves the best performance score of $86.6 \%$.
When DA4LG employs the LGR task exclusively, it yields a performance metric of $81.9 \%$ in validation.
The incorporation of VLC and VGC enhances DA4LG performance, achieving scores of $85.4 \%$ and $85.0 \%$, respectively. 
For different encoders, DA4LG without Vision Encoder achieves $81.5 \%$ in validation and DA4LG without Domain-specific Encoder achieves $82.9 \%$.
The DA4LG equipped with a Vision Encoder and Domain-specific Encoder demonstrates an enhanced performance of $86.8 \%$, compared to its performance with only a single encoder.

\subsubsection{Case Study and Visualization}

\begin{figure*}[t]
\centering
\includegraphics[width=0.7\linewidth]{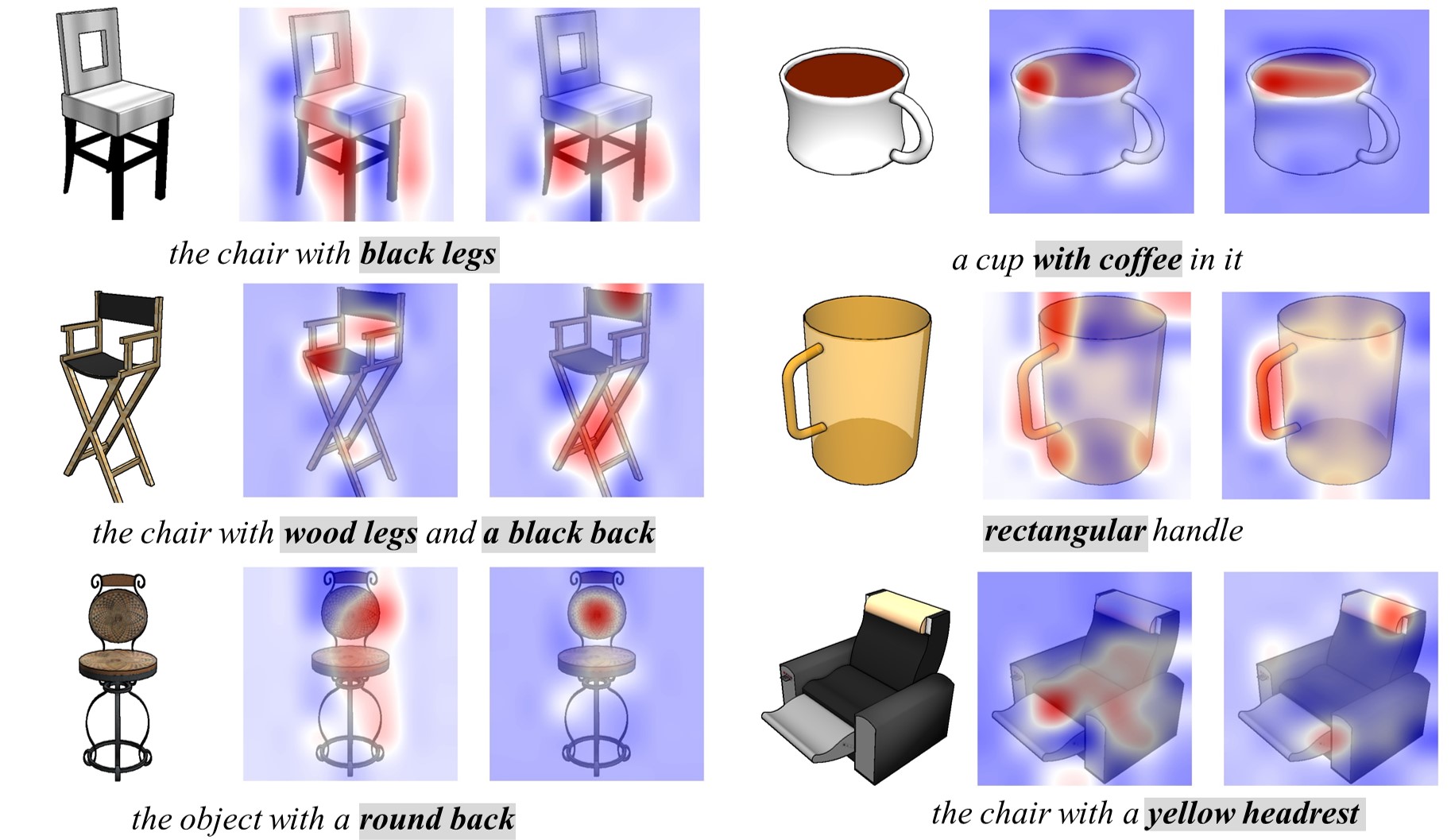}
\caption{Visualization of examples: Original images of the objects are displayed on the left. 
In the middle, attention score maps are visualized, and on the right, attention score maps are enhanced using a domain adapter in a Domain-specific Encoder. 
Warmer colors, such as red, indicate higher attention scores, while cooler colors, such as blue, represent lower attention scores.}
\label{fig:case}
\end{figure*}

As depicted in Fig.~\ref{fig:case}, we randomly select cases for visualization to show the language vision alignment in the target domain.
The left column denotes the raw image of the target objects.
The middle and right images illustrate the attention score map generated by the Domain-specific Encoder without domain adapter and with domain adapter respectively.
For the descriptions \textit{``a cup with coffee in it''} and \textit{``the chair with a yellow headrest''}, the attention map without the domain adapter fails to capture the descriptive elements such as ``coffee'' and ``headrest'' completely. 
The integration of the domain adapter results in an enhancement in the attention directed towards ``coffee'' and ``headrest''.

Additionally, the attention map extends its focus beyond the intended regions in the absence of the domain adapter. 
For the descriptions like ``\textit{the chair with black legs}'', ``\textit{the chair with wooden legs and a black back}'', ``\textit{rectangular handle}'', and ``\textit{the object with a round back}'', the attention is dispersed across the primary structure of the objects. 
However, the incorporation of the domain adapter refines the attention map, directing focus toward the detailed parts.

\subsection{Simulation Results}
\begin{table}[t]
\caption{Results of simulation experiments comparing existing methods with DA4LG. 
*Results are obtained through our reproduction.
}
    \label{tab:simulation_study}
    \centering
    \scalebox{0.80}{
   \begin{tabular}{@{}c|c|ccc|ccc@{}}
\toprule
\multirow{2}{*}{Models}                                 & \multirow{2}{*}{Method} & \multicolumn{3}{c|}{Multi-view}               & \multicolumn{3}{c}{Single-view}               \\ \cmidrule(l){3-8} 
                                                        &                         & Visual        & Blind         & All           & Visual        & Blind         & All           \\ \midrule
CLIP~\cite{radford2021learning}   & External Prior          & 56.8         & 55.0          & 55.4          & 55.0          & 54.8          & 54.9          \\
MATCH~\cite{thomason2022language}* & External Prior          & 61.5 (0.8)         & 60.2 (0.7)         & 60.5 (0.6)         & 59.9 (0.5)          & 59.6 (0.6)          & 59.7 (0.5)         \\
LAGOR~\cite{thomason2022language}* & Multi-view Perception   & 62.5 (0.5)         & 61.0 (0.6)         & 61.3 (0.5)         & 60.7 (0.5)         & 60.1 (0.8)         & 60.2 (0.6)         \\
VLG~\cite{corona2022voxel}*        & External Prior          & 62.1 (0.7)          & 61.2 (0.4)         & 61.4 (0.4)         & 61.4 (0.6)         & 60.3 (0.8)        & 60.5 (0.7)         \\
\textbf{DA4LG (ours)}                  & Domain Adaptation       & \textbf{64.0 (0.5)} & \textbf{63.8 (0.6)} & \textbf{63.9 (0.8)} & \textbf{63.1 (0.7)} & \textbf{62.4 (0.8)} & \textbf{62.6 (0.5)} \\ \bottomrule
\end{tabular}
}
\vspace{-10pt}
\end{table}

To answer \textbf{RQ4}, we conduct a comparative analysis of the performance between existing methods and DA4LG.
All models are trained on the SNARE dataset and subsequently deployed in a zero-shot setting to Simulation-SNARE.
For our experiments, we provide multiple object observations including \{bird, front, left, right, side\} view images in a simulation environment, of which the bird-view image is the input of a single-view experiment.
Compared to SNARE, Simulation-SNARE has the following characteristics:

\begin{enumerate}

\item[] $\bullet$ \textbf{Viewpoint Diversity}. 
Unlike the fixed viewpoint observed in the SNARE dataset, objects in the simulation environment exhibit a diverse viewpoint. 

\item[] $\bullet$ \textbf{Physical Scene}. 
The presence of physical scenes including the background and the worktop in the simulation environment is more similar to the real-world setting.

\item[] $\bullet$ \textbf{Texture Quality.} 
In the simulated environment, there is a richness in geometric details with a deficiency in textural rendering.

\end{enumerate}

Table~\ref{tab:simulation_study} shows the DA4LG architecture exhibits superior generalization and robustness compared to other existing models in the simulation environment. 
Specifically, DA4LG achieves the highest scores of $63.9 \%$ and $62.6 \%$ in multi-view and single-view settings, respectively, outperforming the suboptimal model VLG by $2.5 \%$ and $2.1 \%$ in the respective settings.

\begin{figure*}[!t]
\centering
\includegraphics[width=0.85\linewidth]{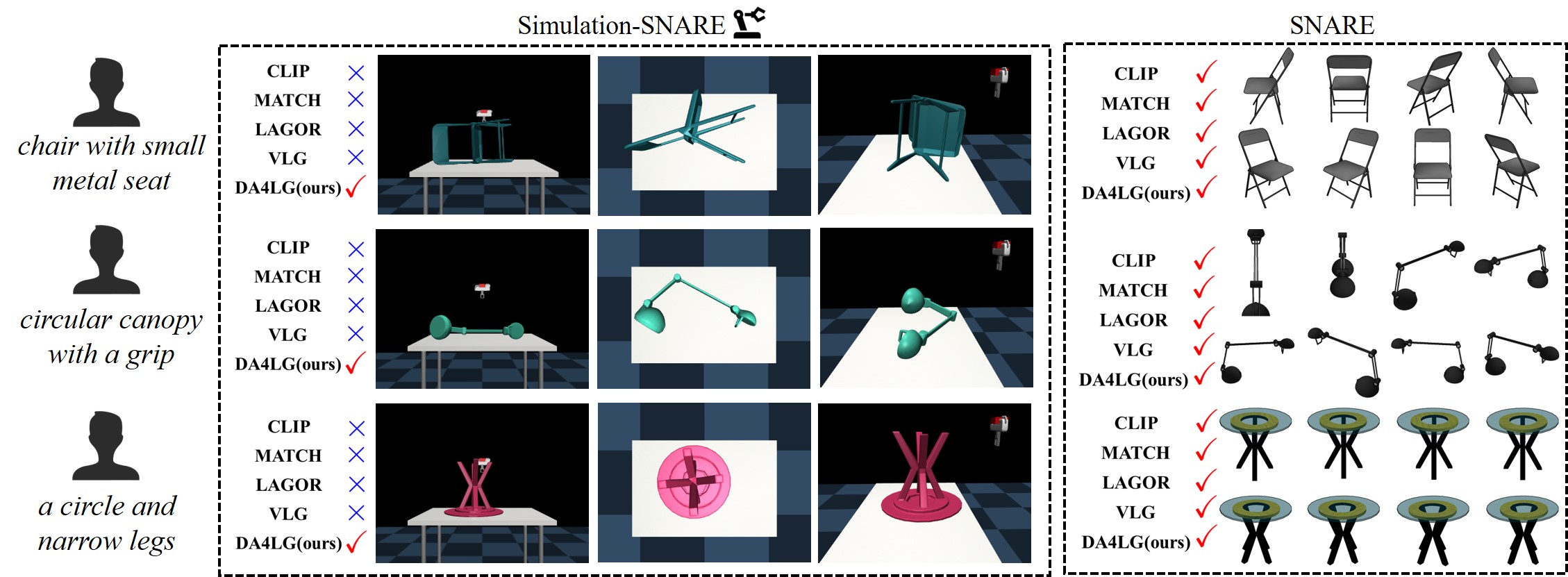}
\caption{ 
Examples illustrating instances where existing methods demonstrate success ($\color{red}\checkmark$) in the SNARE dataset but failure ($\color{blue}\times$) in the Simulation-SNARE dataset, in contrast to DA4LG, which maintains robust performance across both datasets ($\color{red}\checkmark$). 
We visualize the language description (left), Simulation-SNARE (middle), and SNARE (right). For the Simulation-SNARE examples, we showcase the front, bird, and side views.
For the SNARE examples, we showcase all eight views.
}
\label{fig:simulation}
\vspace{-10pt}
\end{figure*}

As illustrated in Figure~\ref{fig:simulation}, We randomly select three cases where other methods (CLIP, MATCH, LAGOR, VLG) succeed in the SNARE dataset but fail in the Simulation-SNARE dataset.
For instance, objects such as a \textit{``chair with a small metal seat''} and a \textit{``circular canopy with a grip''} are depicted in a fallen state, adding complexity to the task.
The object \textit{``a circle and narrow legs''} fails to capture the transparent top of the table in Simulation-SNARE.
The DA4LG model exhibits strong performance in both the SNARE and Simulation-SNARE, indicating its enhanced suitability for simulation environments and operational scenes in the real scene.





\section{Conclusion}
In this work, we attempt to address language grounding tasks from the perspective of domain adaptation and introduce a novel method named \textbf{DA4LG}.
A Domain-specific Encoder and a multi-task learning framework are proposed to improve language-based 3D object understanding, in which aligned cross-modal representation and domain information is encoded effectively. 
Evaluations on the benchmark demonstrate that DA4LG achieves SOTA performance of $83.8 \%$ and $86.8 \%$ in single-view and multi-view settings, respectively. 
The experiment results show the generalization and robustness of our proposed model compared to existing works. 
Our model reduces the domain gap for cross-modal aligned representation.
We also reveal the improvement space in existing methods for domain gap research and underscore the domain adaptation in language grounding with 3D objects.

\vspace{2em}
\noindent \textbf{Acknowledgments}. This work is supported by the Postdoctoral Fellowship Program of CPSF under Grant Number GZC20232292.

%
%

\end{document}